\documentclass[webpdf,modern,medium]{oup-authoring-template}

\onecolumn 

\graphicspath{{Fig/}}

\theoremstyle{thmstyleone}%
\theoremstyle{thmstyletwo}%
\theoremstyle{thmstylethree}%

\usepackage{amsmath}
\usepackage{amssymb}
\usepackage{comment}
\usepackage{multirow}
\usepackage{booktabs}
\usepackage{makecell}
\usepackage{graphicx}
\usepackage{tabularx}
\usepackage{textcomp}
\usepackage{listings}
\usepackage{xcolor}

\definecolor{backcolour}{rgb}{0.98,0.98,0.95}
\newcommand{\listingsttfamily}{\fontfamily{NotoSansMono-TLF}\small}
\DeclareMathOperator*{\argmax}{arg\,max}

\begin{document}


\journaltitle{Journal Title Here}
\DOI{DOI HERE}
\copyrightyear{2022}
\pubyear{2019}
\access{Advance Access Publication Date: Day Month Year}
\appnotes{Paper}

\firstpage{1}


\title[]{Data-Efficient Biomedical In-Context Learning: A Diversity-Enhanced Submodular Perspective}

\author[1]{Jun Wang}
\author[1]{Zaifu Zhan}
\author[3]{Qixin Zhang}
\author[1]{Mingquan Lin}
\author[2]{Meijia Song}
\author[1,$\ast$]{Rui Zhang}


\address[1]{\orgdiv{Division of Computational Health Sciences, Department of Surgery}, \orgname{University of Minnesota}, \orgaddress{\street{516 Delaware St SE, Minneapolis}, \postcode{55455}, \state{MN}, \country{USA}}}
\address[2]{\orgdiv{School of Nursing}, \orgname{University of Minnesota}, \orgaddress{\street{516 Delaware St SE, Minneapolis}, \postcode{55455}, \state{MN}, \country{USA}}}
\address[3]{\orgdiv{College of Computing and Data Science}, \orgname{Nanyang Technological University}, \orgaddress{\street{50 Nanyang Avenue}, \postcode{639798}, \country{Singapore}}}

\corresp[$\ast$]{Corresponding author. \href{email:zhan1386@umn.edu}{zhan1386@umn.edu}}

\received{Date}{0}{Year}
\revised{Date}{0}{Year}
\accepted{Date}{0}{Year}



\abstract{
Recent progress in large language models (LLMs) has leveraged their in-context learning (ICL) abilities to enable quick adaptation to unseen biomedical NLP tasks. By incorporating only a few input-output examples into prompts, LLMs can rapidly perform these new tasks.
While the impact of these demonstrations on LLM performance has been extensively studied, most existing approaches prioritize representativeness over diversity when selecting examples from large corpora.
To address this gap, we propose \textit{Dual-Div}, a diversity-enhanced data-efficient framework for demonstration selection in biomedical ICL. Dual-Div employs a two-stage retrieval and ranking process: First, it identifies a limited set of candidate examples from a corpus by optimizing both representativeness and diversity (with optional annotation for unlabeled data).
Second, it ranks these candidates against test queries to select the most relevant and non-redundant demonstrations. Evaluated on three biomedical NLP tasks (named entity recognition (NER), relation extraction (RE), and text classification (TC)) using LLaMA 3.1 and Qwen 2.5 for inference, along with three retrievers (BGE-Large, BMRetriever, MedCPT), Dual-Div consistently outperforms baselines—achieving up to 5\% higher macro-F1 scores—while demonstrating robustness to prompt permutations and class imbalance. Our findings establish that diversity in initial retrieval is more critical than ranking-stage optimization, and limiting demonstrations to 3–5 examples maximizes performance efficiency.
}


\maketitle

\section{Introduction}


Recent years have witnessed the rapid advancement of large language models (LLMs), exemplified by prominent instances like ChatGPT~\cite{achiam2023gpt}, Llama~\cite{grattafiori2024llama}, and Qwen~\cite{yang2024qwen2} series. These models have significantly enhanced few-shot capabilities~\cite{brown2020language} across numerous natural language processing (NLP) tasks.
Within the biomedical domain, however, a critical challenge persists: the scarcity of high-quality training data. This scarcity arises from two primary factors. Firstly, stringent privacy regulations, such as HIPAA, and patient consent requirements strictly limit access to sensitive patient information. Secondly, rare diseases often suffer from a paucity of structured clinical records and well-defined features~\cite{chen2024rarebench}, demanding robust and generalizable algorithmic solutions~\cite{ullah2024challenges}. Consequently, the few-shot learning paradigm, particularly its ability to perform tasks without task-specific training data, becomes critical~\cite{ge2023few}.

In-Context Learning (ICL)~\cite{dong-etal-2024-survey}, a prominent few-shot technique, offers significant promise for biomedical NLP. ICL leverages task-specific prompts containing a few annotated examples, enabling pre-trained LLMs to perform unseen tasks without parameter updates. This approach typically involves retrieving a small set of relevant examples from a large unlabeled corpus and annotating them according to task requirements. ICL confers several key advantages for biomedical NLP.
Firstly, it simplifies the deployment of pretrained LLMs by replacing computationally intensive fine-tuning with prompt design.
More importantly, since a task-specific fine-tuning dataset is no longer required, ICL also reduces the need for labeled data for downstream tasks. By avoiding parameter updates, ICL can potentially improve model stability across diverse tasks.

Crucially, the performance of ICL is highly dependent on the quality of the selected demonstrations. Effective demonstrations provide essential context, supplementing label information and relationships beyond what was learned during pre-training~\cite{kossen2024context}. This is especially vital in biomedicine, where texts are dense with specialized terminology (e.g., genes, cells, pathways). Since domain-specific knowledge cannot be easily integrated post-hoc into general LLMs, carefully curated demonstrations become the primary mechanism for activating or supplementing the relevant internal knowledge encoded during pre-training.



A significant challenge for existing ICL frameworks lies in selecting optimal demonstrations. Traditional methods often rely on a single metric, typically prioritizing examples most representative of or semantically similar to the test query.
More recent approaches~\cite{kumari-etal-2024-end} have adopted active learning principles to select groups that maximally cover the semantic space of the corpus. However, a key limitation persists: the neglect of diversity among the selected examples. Providing diverse demonstrations enhances model robustness by exposing it to varied scenarios. In biomedicine, this diversity is critical not only for performance but also for improving fairness in disease representation and supporting more comprehensive clinical decision-making.

To address this gap, we propose Dual-Div, a novel diversity-enhanced ICL framework for biomedical NLP that explicitly optimizes demonstration selection across multiple dimensions—representativeness and diversity. Dual-Div operates in two stages, i.e., demonstration retrieval and ranking. 
In the first stage, we recall a limited set of candidate examples from a large (often unlabeled) corpus, followed by annotation if necessary.
In the second stage, these candidates are ranked in conjunction with test queries, selecting the top demonstrations for prompt construction.
Dual-Div provides a systematic and effective solution for balancing broad semantic coverage with intrinsic diversity within the selected demonstrations. Our comprehensive evaluations demonstrate that this approach yields significant improvements in ICL performance over prior methods.
Our key contributions are summarized as follows.
\begin{itemize}
    \item To the best of our knowledge, this is the first study to systematically integrate diversity metrics into the demonstration selection process for in-context learning.
    
    \item We propose a novel two-stage submodular optimization framework that jointly maximizes semantic coverage and diversity of selected demonstrations.

    \item We conduct extensive experiments, evaluating the framework across 2 LLMs and 3 retrievers on 3 datasets, demonstrating state-of-the-art results and providing comprehensive insights.
\end{itemize}

\section{Related Work}

As model and data scales increase, ICL emerges as a distinctive capability of LLMs~\cite{brown2020language, dong-etal-2024-survey}. This paradigm enables LLMs to perform unseen biomedical tasks by learning directly from demonstration examples provided within their input context.
Mirroring human analogical reasoning, ICL offers an interpretable interaction mechanism: it explicitly activates domain-specific knowledge encoded within LLM parameters through natural language demonstrations, bypassing traditional parameter updates.
This section reviews relevant literature through two lenses: (1) the application of LLMs in biomedicine, and (2) the critical challenge of demonstration selection for effective ICL.


\subsection{Biomedical applications of LLMs}

Evolving from earlier transformer-based architectures~\cite{lee2020biobert}, LLMs have become fundamental NLP tools demonstrating significant potential across biomedical informatics~\cite{tian2024opportunities} and healthcare~\cite{karabacak2023embracing}.
In text mining, LLMs drive substantial improvements in core tasks including medical text summarization, information extraction, question answering, and medical education, often surpassing previous state-of-the-art results~\cite{yang2022large, singhal2025toward, safranek2023role}. Beyond these, promising applications extend into health-related domains. LLMs show utility in providing valuable biomedical insights~\cite{shool2025systematic}, automating clinical coding~\cite{soroush2024large}, and aiding interpretable differential diagnosis (DDx)~\cite{zhou2024interpretable, zhou2025large}.
Furthermore, they demonstrate auxiliary benefits in downstream tasks like drug discovery and repurposing~\cite{zheng2024large, xiao2024repurposing}.

Notably, LLMs can achieve performance comparable to human experts in specific domains while offering potentially interpretable reasoning to support comprehensive decision-making. Nevertheless, critical challenges—including mitigating biases, ensuring data security, and addressing ethical concerns—remain urgent barriers to the broader adoption of biomedical LLMs.





\subsection{Demonstration Selection}

Unlike retrieval-augmented generation (RAG)~\cite{lewis2020retrieval}, which dynamically retrieves external, up-to-date knowledge relevant to a query, ICL primarily queries the LLM's internal knowledge. Its efficacy hinges on the model's ability to infer task requirements solely from the provided context demonstrations. Consequently, ICL performance is highly sensitive to the quality, relevance, and composition of these selected demonstrations.


Crucially, while ICL was originally conceived as a training-free method to reduce computational cost, many current approaches incur significant auxiliary costs during demonstration selection.
For instance, some methods require data preparation and supervised training to learn specialized retrievers~\cite{rubin-etal-2022-learning, ye2023compositional}, while others leverage the LLM's inference capabilities during selection (e.g., using confidence scores)~\cite{su2023selective}, adding latency and compute overhead.
This paper specifically focuses on learning-free, end-to-end ICL frameworks. We prioritize methods where the LLM itself is not utilized within the demonstration selection algorithm; its role is confined to the final inference step after demonstrations are fully determined.


Retrieving effective demonstrations from large corpora (labeled or unlabeled) often starts with simple heuristics like cosine similarity, which remains surprisingly effective and widely used~\cite{wu2023self, margatina2023active, zhan2025mmrag}.
Moving beyond basic similarity, research has explored optimization objectives like the facility location function~\cite{lin2009graph}, which emphasizes the coverage or representativeness of the selected set. While Div-S3~\cite{kumari-etal-2024-end} offers a theoretically grounded framework using this function, it primarily addresses coverage and neglects diversity. The more recent diversity-guided search~\cite{li2023finding} addresses this gap by incorporating heuristic beam search to promote diverse examples.
Finally, it is essential to note that once demonstrations are selected, factors like prompt formatting and the permutation order of examples within the context significantly influence the accuracy of the LLM's response~\cite{zhao2021calibrate, liu2022makes, lu2022fantastically}.



\section{Problem Formulation}

ICL is a kind of few-shot learning method enabling LLMs to adapt to new tasks with only a small number of task-specific demonstration examples~\cite{brown2020language}.
Let $V$ denote the corpus, with or without task-specific labels, composed of $N=|V|$ datapoints $(x_i, y_i, \mathbf{e}_i)_{i=1}^N$, where $x_i$ is the input sentence, $y_i$ is the label, and $\mathbf{e}_i$ is the representation vector from a retriever model.
Here, we denote this retriever model by $g(\cdot)$, which means $\mathbf{e}_i = g(x_i)$.

Given a test query set $Q$ that includes samples $(x_{\text{test}}, y_{\text{test}})$, the target is to select a subset $T^* \subseteq V$ such that the likelihood of the target answer $y_{\text{test}}$ from the scoring distribution outputted by LLMs, denoted by $\Pi_{(x_{\text{test}}, y_{\text{test}}) \in Q} P(y_{\text{test}} \mid x_{\text{test}})$, is maximized.
During this process, a major destination is to limit the cardinality of $T^*$, i.e., ensuring that $|T^*| \leq k$ with $k << N$.
Equivalently, this constraint can also be considered as the maximal sequence length of the prompt instruction~\cite{lin2010multi}.

\begin{figure}[ht]
    \centering
    \includegraphics[width=0.95\columnwidth]{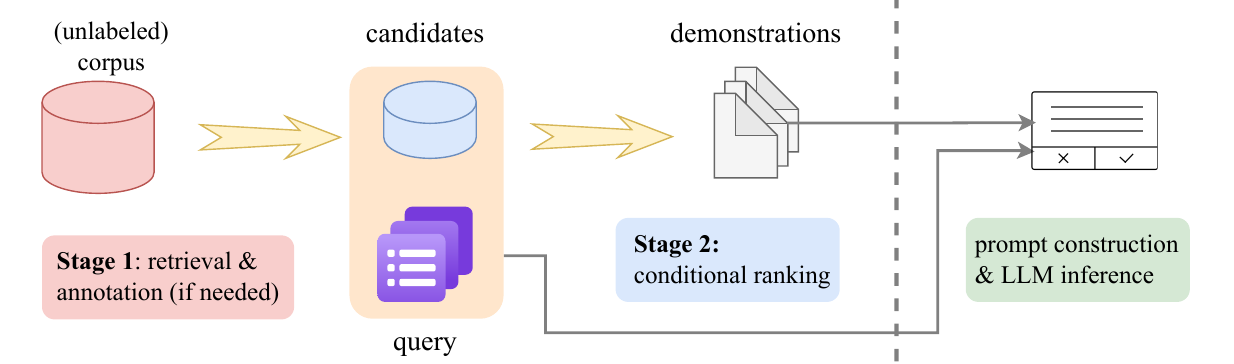}
    \caption{The workflow of Dual-Div, a two-stage in-context learning framework.
    }
    \label{fig_icl_flow}
\end{figure}

In general, this process can be further split into two procedures, as illustrated in Figure~\ref{fig_icl_flow}. For the first one, a limited number (namely $k_1$, $k < k_1 << N$) of candidate examples, denoted by $S^*$, are retrieved from the corpus $V$, in terms of the \textbf{coverage} and \textbf{diversity}.
Intuitively, what we expect is to select a subset $S$ that contains the most relevant and non-redundant examples from the corpus set $V$. As a result, the two metrics for coverage and diversity are critical. Let $\mathcal{C}(S)$ and $\mathcal{D}(S)$ be the corresponding notations. The final utility function that is composed of these two perspectives can be written as
\begin{equation}
    f(S) = \mathcal{C}(S) + \lambda \mathcal{D}(S),
\end{equation}
where $\lambda$ is a trade-off parameter.
With $f(S)$ as our optimization objective, we can conclude the first procedure as
\begin{equation}\label{problem_first}
    S^* = \argmax_{S \subseteq V, |S| \leq k_1} f(S).
\end{equation}
If the samples in $S^*$ are unlabeled, an extra annotation step is required. However, this approach has already reduced costs by avoiding annotation of the entire corpus $V$.

In the second stage, given the test query set $Q$, the informative examples in the subset $S^*$ are ranked using the same metrics — now applied conditionally as $f(T \mid Q)$ with $T \subseteq S^*$.
The optimization problem associated with this step can be written as
\begin{equation}
    T^* = \argmax_{T \subseteq S^*, |T| \leq k} f(T \mid Q).
\end{equation}

Finally, the selected demonstrations in $T^*$ and the test query $x_{\text{test}}$ in $Q$ will be used to construct the prompt for LLM inference.
The evaluation is based on the likelihood
\begin{equation}
    \Pi_{(x_{\text{test}}, y_{\text{test}}) \in Q} P(y_{\text{test}}\mid x_{\text{test}}) = \Pi_{(x_{\text{test}}, y_{\text{test}}) \in Q} f(y_{\text{test}} \mid \mathcal{T}(T^*, x_{\text{test}})),
\end{equation}
where $\mathcal{T}$ denotes the instruction prompt template regarding the examples, test query, and task descriptions.

Again, our proposed framework aims to address two important issues associated with efficient in-context learning:

(1) \textit{Given a large corpus $V$, how can we identify a subset with cardinality constraint to keep the most information while ensuring the internal diversity at the same time?}

(2) \textit{Given a query $Q$, how can we decide the most relevant and non-redundant demonstrations from $S^*$?}

\section{The Proposed Framework}

In this section, after introducing our choice for the metric of diversity and coverage, we will explain the technical details of the Dual-Div framework.

\subsection{Metric Selection}

Ideally, we desire to find suitable monotone submodular functions to describe diversity and coverage such that the weighted form $f(S)$ still belongs to monotone submodular functions. With this assumption, we can obtain a theoretically guaranteed optimal solution.

\noindent \textbf{Definition of Diversity.} To seek an elegant probabilistic model to describe the probabilities of subsets $S$, we leverage the algebraic properties of determinantal point processes (DPPs)~\cite{borodin2009determinantal}.
Specifically, we select the L-ensemble representation of DPPs with a semidefinite matrix $\mathbf{L}$. Given a subset $S \subseteq V$, the probability that it would be selected is
\begin{equation}
    \mathcal{P}(S) = \frac{\text{det}(\mathbf{L}_{S})}{\text{det}(\mathbf{L}+ \mathbf{I})}
\end{equation}
where $\mathbf{L}_{S}=\left[\mathbf{L}_{ij}\right]_{i,j \in S}$ denotes the restriction of $\mathbf{L}$ to the entries indexed by elements in $S$, $\text{det}(\cdot)$ denotes the determinant of a matrix, and $\mathbf{I}$ represents the identity matrix. Note that $\mathcal{P}(S)$ is normalized because of the equation $\sum_{S \subseteq V} \text{det}(\mathbf{L}_{S})=\text{det}(\mathbf{L}+\mathbf{I})$.

Similar to other scenarios like recommendation~\cite{kulesza2012determinantal, chen2018fast}, we can construct the kernel matrix in the context of ICL by the Gram matrix $\mathbf{L}=\mathbf{E}^T \mathbf{E}$, where the $i$-th columns of $\mathbf{E}$ corresponds the vector $\mathbf{e}_i$ representing the query $i$ in $V$.
Intuitively, the determinant of the gram matrix $\mathbf{L}$ describes the ``volume'' of the space spanned by all query representations in $\mathbf{E}$.
Therefore, if the queries in $S$ are more various, the spanned space will be greater, resulting in a greater determinant of $\mathbf{L}_{S}$.
Furthermore, if we rewrite $\mathbf{e}_i$ as the product of the $L_2$ norm $r_i \geq 0$ and a normalized vector $\bar{\mathbf{e}}_i$ with $||\bar{\mathbf{e}}_i||_2=1$.
The elements of kernel $\mathbf{L}$ can be written as
\begin{equation}
    \mathbf{L}_{i j}
    =\left\langle \mathbf{E}_i, \mathbf{E}_j\right\rangle
    =\left\langle r_i \bar{\mathbf{e}}_i, r_j \bar{\mathbf{e}}_j \right\rangle
    =r_i r_j\left\langle\bar{\mathbf{e}}_i, \bar{\mathbf{e}}_j\right\rangle
    =r_i r_j w_{i,j}.
\end{equation}
Here, the term $w_{i,j}$ is exactly the cosine similarity of query $i$ and $j$.
Due to the equation $
\mathbf{L}=\text{Diag}\left(\mathbf{r}\right) \cdot \mathbf{W} \cdot \text{Diag}\left(\mathbf{r}\right)$, the problem of maximizing $\mathcal{P}(S)$ can be thought of as 
\begin{equation}\label{eq_log_det}
    \max_{S \subseteq V} \log \mathcal{P}(S) = \max_{S \subseteq V} \log \text{det}\left(\mathbf{L}_{S}\right)
    =\max_{S \subseteq V} \left[\sum_{i \in S} \log \left(r_i^2\right) + \log \text{det} \left(\mathbf{W}_{S}\right) \right].
\end{equation}

This decomposition reveals that the probabilities of DPP models can be further split into a diversity-related factor $\log \text{det} \left(\mathbf{W}_{S}\right)$ and a norm-related factor $\sum_{i \in S} \log \left(r_i^2\right)$.
Thus, we propose to define the diversity of the subset $S$ without the influence of the $L_2$ norm of query embeddings, i.e.,
\begin{equation}
    \mathcal{D}(S)=\log \text{det} \left(\mathbf{W}_{S}\right),
\end{equation}
which also aligns with the definition of $\mathcal{C}(S)$ below.




\noindent \textbf{Definition of Coverage.} Following the previous studies~\cite{lin2009graph, kumari-etal-2024-end}, we then define the coverage of a subset $S$ regarding the entire set $V$ as the facility location function
\begin{equation}\label{eq_coverage}
    \mathcal{C}(S)=\sum_{i \in V} \max_{j \in S} w_{i,j}
    =\sum_{i \in V} \max_{j \in S} \left\langle\bar{\mathbf{e}}_i, \bar{\mathbf{e}}_j\right\rangle.
\end{equation}

As such, the ultimate objective is a balance of diversity and coverage metric as
\begin{equation}\label{eq_f_S}
    f(S) = \mathcal{C}(S) + \lambda \mathcal{D}(S)
    =\sum_{i \in V} \max_{j \in S} w_{i,j} + \lambda \log \text{det} \left(\mathbf{W}_{S}\right),
\end{equation}
where $\left[\mathbf{W}_{S}\right]_{ij}=w_{i,j}$ for $i, j \in S$.
We can prove that $f(S)$ is monotone and submodular for a small trade-off constant $\lambda$ (See Appendix~\ref{app_proof}).

\subsection{Diversity-Enhanced Retrieval}

With the utility function $f(S)$ at hand, we first compute a cosine similarity matrix for the entire set $V$ and use it to instantiate the value of $f(S)$ for every single-element set $S=\{x_i\}_{x_i \in V}$.
Then, we add one sample with the largest marginal gain $f(S \cup \{x\}) - f(S)$ each time until the cardinality of $S$ reaches the limit $k_1$. During this process, the lazy greedy~\cite{minoux2005accelerated} strategy is applied to accelerate the efficiency.

Since we only consider the special cardinality constraint so far, where the costs of all elements are identical, it has been shown in \cite{nemhauser1978analysis} that the greedy optimization provides a $(1-1/e)$-approximation guarantee.
In other words, we can make sure the $f(S^*) \geq (1-1/e)f(S_{\text{opt}})$ where $S_{\text{opt}}$ denotes the ideal optimal subset.
For a more general case where the constraint $|S|\leq k_1$ in Eq.~\eqref{problem_first} is replaced with $\sum_{x \in S} c(s) \leq k_1$, a modified greedy algorithm proposed in \cite{lin2011class} alternatively offer a $(1-1/\sqrt{e})$ performance guarantee factor.

        

\subsection{Diversity-Enhanced Ranking}
Once we have retrieved a subset $S$ that covers as many semantic meanings of $V$ as possible, we now consider the conditional ranking process given the test query set $Q$.
At a high level, we expect to generate a subset with the largest marginal advantage given the query set $Q$, where the marginal advantage of a subset $T \subseteq S^*$ is denoted by
\begin{equation}
    f(T \mid Q) := f(T \cup Q) - f(Q).
\end{equation}
A higher value of $f(T \mid Q)$ indicates a higher utility gain if we combine the samples in $T$ with those in $Q$, where the utility depends on our objective, such as the relevance or non-redundancy.
Here, we use the same definition of $f(T)$ as used in the previous stage -- a balance of coverage and diversity.

In practice, we utilize a modular approximation $\sum_{x_i \in T} f\left( \{x_i\} \mid Q\right)$ as the upper bound of $f(T \mid Q)$, resulting the final objective of this stage to be
\begin{equation}\begin{aligned}
    T^* &= \argmax_{T \subseteq S^*, |T| \leq k} \sum_{x_i \in T} f( \{x_i\} \mid Q) \\
    &=\argmax_{T \subseteq S^*, |T| \leq k} \sum_{x_i \in T} f( \{x_i\} \cup Q) - f(Q).
\end{aligned}\end{equation}
We also conclude the detailed algorithms for the entire process, including the retrieval and ranking in Algorithm~\ref{algo_dualdiv} in Appendix~\ref{sec_appendix_algo}.
\section{Experiments}
In this section, after introducing the detailed experimental settings, we present the results from a comprehensive perspective.

\subsection{Datasets}
We evaluate on three core biomedical NLP tasks, including named entity recognition (NER), relation extraction (RE), and text classification (TC).
For each task, we select one typical dataset -- ChemProt for NER, DDI for RE, and HealthAdvice for TC.

\begin{itemize}

    \item \textbf{ChemProt}~\cite{krallinger2017overview}
    This dataset consists of 1,820 PubMed abstracts with chemical-protein interactions annotated by domain experts. Therefore, it can also be used to evaluate the systems that are able to detect the biomedical terminology related to chemical compounds/drugs and genes/proteins.

    \item \textbf{DDI}~\cite{segura2013semeval}. 
    This dataset is identified as a typical biomedical relation extraction benchmark to evaluate the ability of extracting drug-drug interactions.
    The potential types includes \textit{advice}, \textit{effect}, \textit{mechanism}, and \textit{int}.
    All drug entities and interactions in sentences are annotated from biomedical literature and drug product information sources.

    \item \textbf{HealthAdvice}~\cite{yu2019EMNLPCausalLanguage}. This dataset is a diverse collection of health-related advice, annotated for text classification tasks related to health information and advisory content.
    Structured to facilitate automatic classification of health advice into relevant categories, it can support applications such as misinformation detection, personalized health recommendations, and automated triaging of medical inquiries.
\end{itemize}

These tasks can comprehensively demonstrate the ability of LLMs in biomedical natural language understanding and clinical decision support.

\subsection{Models}

The involved language models cover two categories. The smaller retriever models are utilized to quickly obtain the semantic representations of biomedical resources.
With these semantic vectors, we can determine the demonstration examples in natural language and derive the final prompts for LLMs.
In contrast, the LLMs, with many more parameters, are more powerful in solving the concrete downstream tasks.

For retriever models, we select three different ones, including BMRetriever~\cite{xu2024bmretriever}, MedCPT~\cite{jin2023medcpt}, and BGE-Large~\cite{xiao2024c}. Among them, BMRetriever and MedCPT are trained on biomedical resources, whereas BGE-Large is a family of embedding models trained on general data.
For the inference LLMs, we select the representative Qwen2.5-7B~\cite{yang2024qwen2} and Llmama 3.1-8B~\cite{grattafiori2024llama} to balance the performance and efficiency during evaluation.

\subsection{Baselines}

For the baseline methods, we adopt the heuristic \textit{Random-Similar} algorithm in addition to the most recent \textit{Div-S3}~\cite{kumari-etal-2024-end} method.
Similar to our Dual-Div, these methods are all two-stage ones. For the first stage, random means randomly selecting examples from the entire set $V$, whereas Div means $f(S)$ is exactly $\mathcal{C}(S)$ without the consideration of $\mathcal{D}(S)$.
For the second stage, similar means selecting the most similar examples conditioned with test set $Q$ (with the highest average cosine similarity), whereas the S3 degrades the utility from $f(T \mid Q)$ to $\mathcal{C}(T \mid Q)$.
To clearly present the ablation influence of the diversity metric $\mathcal{D}(S)$ and $\mathcal{D}(T \mid Q)$, we also introduce two variants of Div-S3, which is named Div*-S3 and Div-S3*.

\subsection{Metrics}

We note that the label distributions in the selected datasets are highly imbalanced. For instance, in the DDI dataset, only 979 out of 5,761 test queries contain one of the four target relations (effect, advice, mechanism, or int), whereas this ratio is 23.6\% for the HealthAdvice dataset.
Given this imbalance, we employ macro-F1 score—in addition to accuracy—to better reflect in-context learning performance across tasks. Unlike micro-F1, macro-F1 equally weights all classes, making it more rational for biomedical applications where rare classes are often as critical as frequent ones~\cite{hassan2024optimizing}.

For the remaining implementation details, we supplement them in Appendix~\ref{sec_appendix_details}. In particular, we list the prompt template we utilize in Appendix~\ref{sec_appendix_template}.

\section{Results}

\subsection{ICL Performance}

\begin{table}[ht]
    \centering
    \caption{ICL Performance on different NLP datasets using Llama 3.1 (8B) and Qwen 2.5 (7B) as inference LLMs and BGE-Large as the retriever.}\label{tab_main_bge}
    \begin{tabular}{clcccccc}
    
    \toprule
    \multirow{2}{*}{LLM} & \multicolumn{1}{c}{\multirow{2}{*}{Method}} & \multicolumn{2}{c}{ChemProt (NER)} & \multicolumn{2}{c}{DDI (RE)} & \multicolumn{2}{c}{HealthAdvice (TC)} \\
    \cmidrule(lr){3-4} \cmidrule(lr){5-6} \cmidrule(lr){7-8}
    & & acc. & macro-F1 & acc. & macro-F1 & acc. & macro-F1 \\
    \midrule
    \multirow{5}{*}{Qwen 2.5} & Random-Simliar & 0.6488 & 0.6764 & 0.7837 & 0.2515 & 0.7592 & 0.3306 \\
    & Div-S3 & 0.7585 & 0.7067 & 0.7004 & 0.2843 & \textbf{0.7661} & 0.3330 \\
    & Div*-S3 & \textbf{0.7691} & 0.7029 & \textbf{0.8047} & 0.2794 & 0.7615 & 0.3604 \\
    & Div-S3* & 0.7212 & 0.7004 & 0.7086 & 0.2849 & 0.7598 & 0.3623 \\
    & Dual-Div & 0.7598 & \textbf{0.7117} & 0.7477 & \textbf{0.2949} & 0.7642 & \textbf{0.3686} \\
    \midrule
    \multirow{5}{*}{Llama 3.1} & Random-Simliar & 0.6046 & 0.5819 & 0.6888 & 0.2646 & 0.6832 & 0.2780 \\
    & Div-S3 & \textbf{0.6921} & 0.6026 & 0.7301 & 0.2653 & 0.7160 & 0.2892 \\
    & Div*-S3 & 0.6912 & 0.6345 & \textbf{0.7351} & 0.3086 & \textbf{0.7586} & 0.2886 \\
    & Div-S3* & 0.6558 & 0.6183 & 0.7256 & 0.2664 & 0.6872 & 0.2875 \\
    & Dual-Div & 0.6910 & \textbf{0.6387} & 0.7283 & \textbf{0.3129} & 0.7241 & \textbf{0.2948} \\
    \bottomrule
    \end{tabular}
\end{table}

Table~\ref{tab_main_bge} presents the in-context learning performance (accuracy and macro-F1) across different tasks and methods.
Utilizing the BGE-large retriever model, Dual-Div achieves the highest macro-F1 score regardless of whether Qwen2.5 or Llama serves as the inference LLM.
This result indicates that the demonstration examples selected by Dual-Div effectively enhance the inference capability of LLMs on biomedical resources.
Notably, Dual-Div also maintains a strong balance between accuracy and macro-F1 performance. This balance likely stems from our method's trade-off between coverage and diversity when selecting demonstrations.
Given the significant class imbalance inherent in biomedical queries (where instances with meaningful relations are vastly outnumbered by non-biomedical ones), we argue that macro-F1 here serves as a more reliable and persuasive evaluation metric than accuracy.
Results for the other two retriever models are provided in Appendix~\ref{sec_appendix_main_exp}; overall, these findings align with our observations using the BGE-Large retriever.


For the comparison of the two LLMs, Qwen2.5 significantly outperforms Llama 3.1 on NER and TC tasks, regardless of the retrievers used. This suggests Qwen2.5 may have a stronger potential for transferring biomedical knowledge from external inputs.
In practice, we also observed more formatting errors in Llama 3.1's outputs than in Qwen2.5's. For example, Llama 3.1 responses for relation types sometimes include (partial) parentheses. This issue diminishes as the number of fine-tuning steps over the corpus $V$ increases.

Another critical result stems from our detailed comparisons of introducing the diversity term $\mathcal{D}(S)$ separately in the first and second stages, corresponding to the performance of Div*-S3 and Div-S3*, respectively.
Crucially, incorporating the diversity regularization term $\mathcal{D}(S)$ in the first stage leads to significant gains in the ICL performance of LLMs, often achieving the best accuracy results.
This suggests that retrieving highly diverse candidate demonstrations from the original corpus $V$ in Stage 1 plays a more decisive role in the final outcomes than employing a sophisticated ranking algorithm in Stage 2. This is particularly true when the candidate pool reduction in Stage 1 ($N - k_1$) is larger than the final selection reduction in Stage 2 ($k_1 - k$).


\subsection{Visualization of diversity}



\begin{figure}[ht]
    \centering
    \includegraphics[width=0.95\columnwidth]{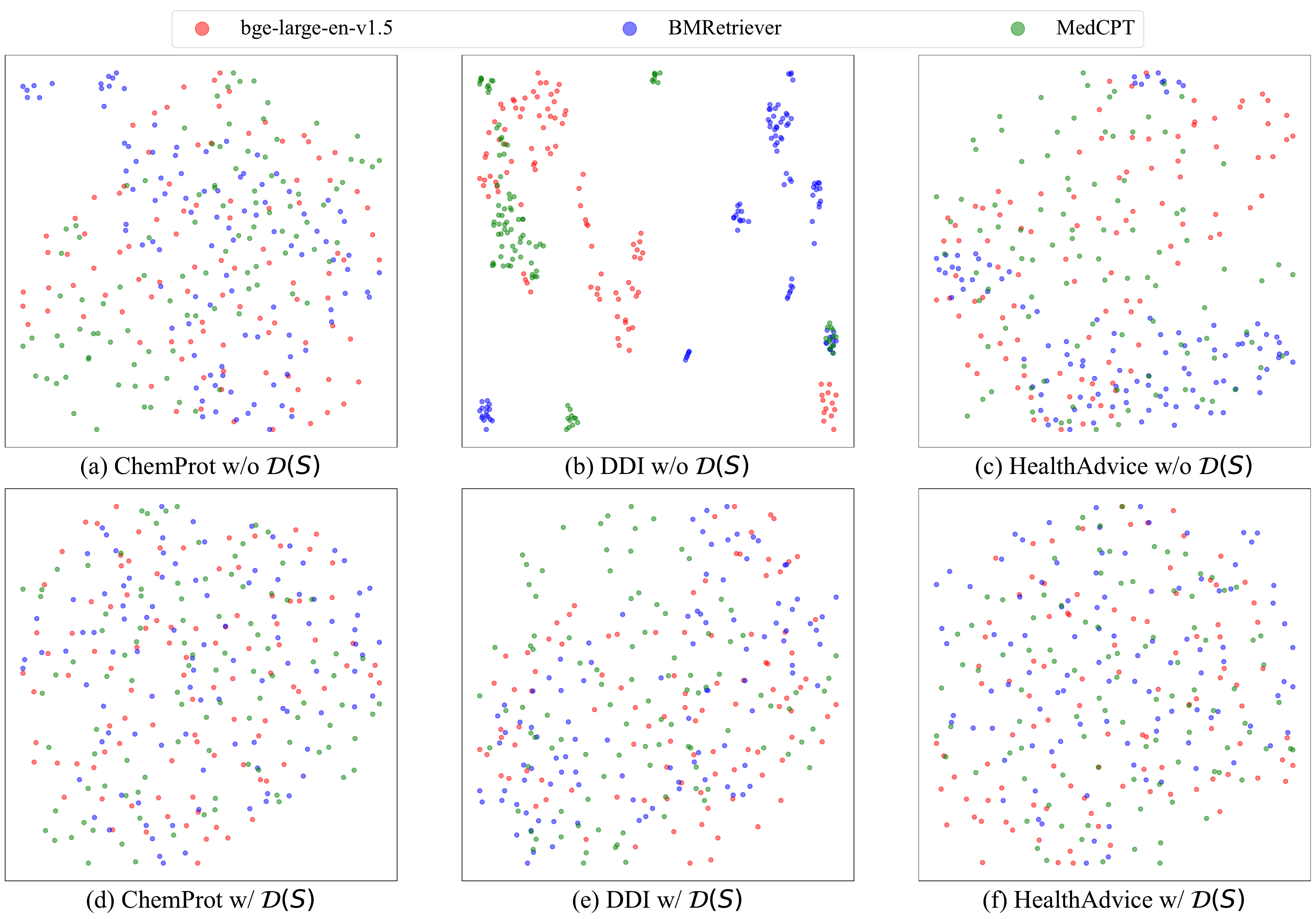}
    \caption{Visualization of retrieved queries in $S^*$ after Stage 1. Semantic embedding vectors from various retriever models w/o and w/ diversity term $\mathcal{D}(S)$ during optimization are reduced and projected by UMAP.}
    \label{fig_visualize}
\end{figure}

To better illustrate the effect of the diversity-enhanced technique, we also visualize the semantic representations of the 100 candidate queries from the first retrieval stage.
Specifically, we use UMAP~\cite{mcinnes2018umap} to project the high-dimensional semantic embedding vectors from the retriever models onto a 2D plane. We then apply MinMaxScaler for normalization to facilitate visualization.
The results, summarized in Fig.~\ref{fig_visualize}, show that queries retrieved using the diversity metric exhibit greater dispersion in their vector representations compared to those retrieved without it.
This difference is particularly pronounced on the DDI dataset for the relation extraction task.
This outcome aligns with our intuition of $\mathcal{D}(S)$ regarding the determinant of the Gram matrix of these vectors. By maximizing the ``volume'' of the space spanned by the semantic vectors, we encourage the model to avoid selecting queries clustered in a small neighborhood, which are likely to represent the same biomedical terminology or phenomenon.

\subsection{Sensitivity Analysis}

In order to investigate the sensitivity of our Dual-Div algorithm, we mainly study the effect of internal ordering of final demonstration sets, as well as the value of the budget constraint $k$, which limits the number of examples in the prompts.

\begin{figure}[ht]
    \centering
    \includegraphics[width=0.98\columnwidth]{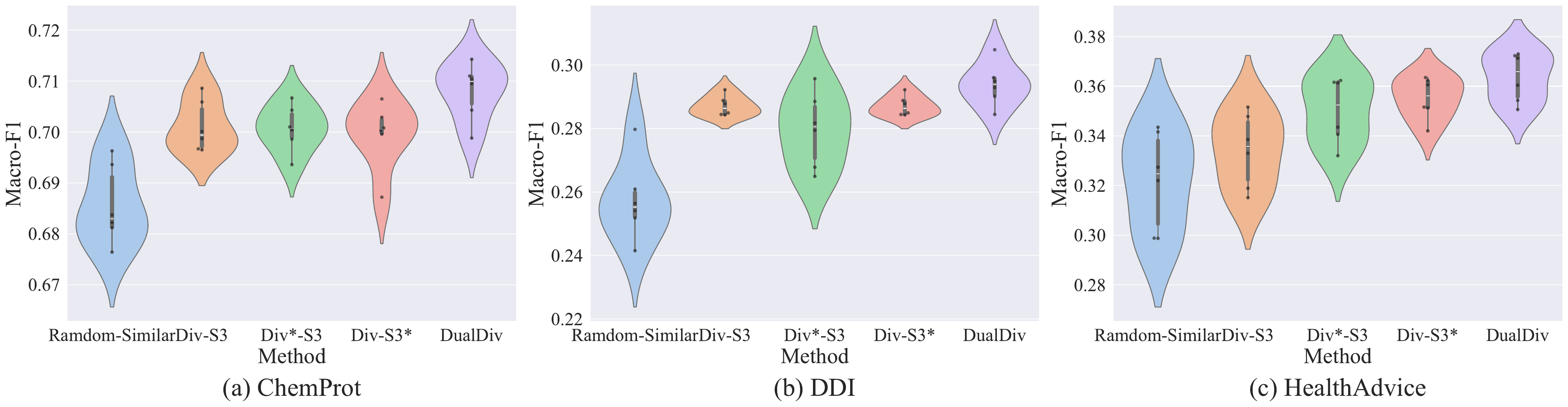}
    \caption{Sensitivity analysis regarding the permutation ordering in the LLM's prompts of the final demonstration set. Given BGE-Large as the retriever model, for each set of 3 demonstrations from different methods, we plot the violin graphics of macro-F1 scores of all 6 possible permutations.}
    \label{fig_permute_bge}
\end{figure}

\subsubsection{On the Effect of Example Ordering}
Previous studies~\cite{zhao2021calibrate, liu2022makes, lu2022fantastically} have shown that the ICL performance of LLMs is highly sensitive to the ordering of demonstrations within input prompts.
To investigate this effect, we evaluate the performance distributions across all possible permutations of the examples in the final set $T$ (default cardinality 3). Fig.~\ref{fig_permute_bge} presents the results for our Dual-Div method and all baselines, using BGE-Large as the retriever. Results for other retriever models are provided in Appendix~\ref{sec_appendix_order}.

Overall, Dual-Div exhibits the strongest robustness against prompt permutations. On ChemProt and HealthAdvice, Dual-Div not only achieves the highest median macro-F1 scores but also shows the tightest performance distribution (narrowest violin shape), indicating minimal sensitivity to permutation order.
Conversely, Random-Similar displays the broadest performance spread across all datasets, likely due to the inherent randomness introduced in its first stage.

On the DDI dataset, Dual-Div shows a much wider distribution than Div-S3 and Div-S3*, but achieves superior average performance. This suggests that prioritizing high diversity in candidate queries during the first stage may reduce stability when using BGE-Large. However, we posit that this effect is both task-specific and retriever-dependent. Supplemental experiments confirm that incorporating diversity significantly improves stability intervals: for ChemProt using BMRetriever (Fig.~\ref{fig_permute_bm}a) and for DDI using MedCPT (Fig.~\ref{fig_permute_medcpt}b).


\begin{figure}[ht]
    \centering
    \includegraphics[width=0.98\columnwidth]{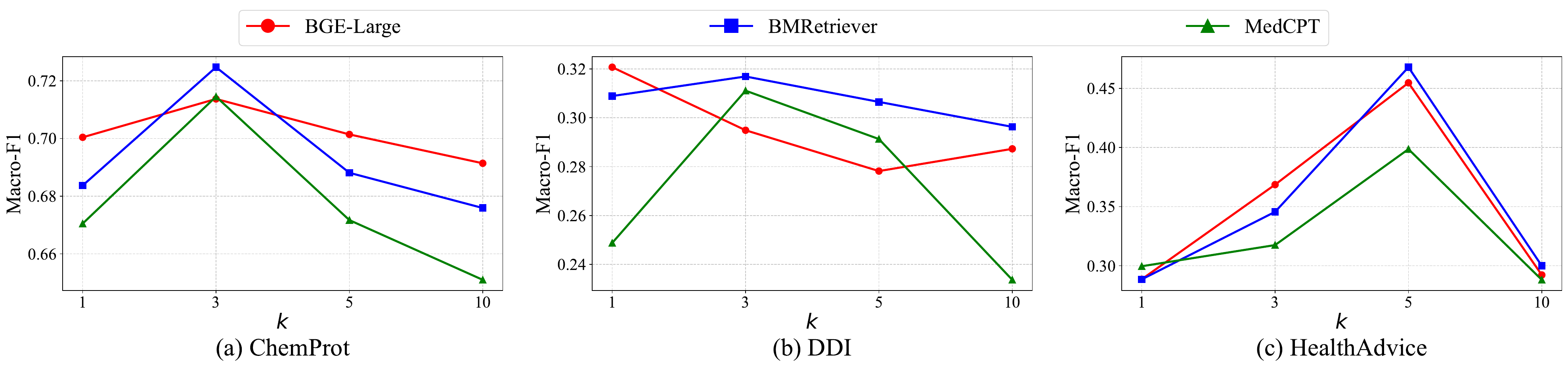}
    \caption{Sensitivity analysis of the budget constraint in the second stage, with Qwen 2.5 as the inference LLM.}
    \label{fig_num_qwen}
\end{figure}

\begin{figure}[ht]
    \centering
    \includegraphics[width=0.98\columnwidth]{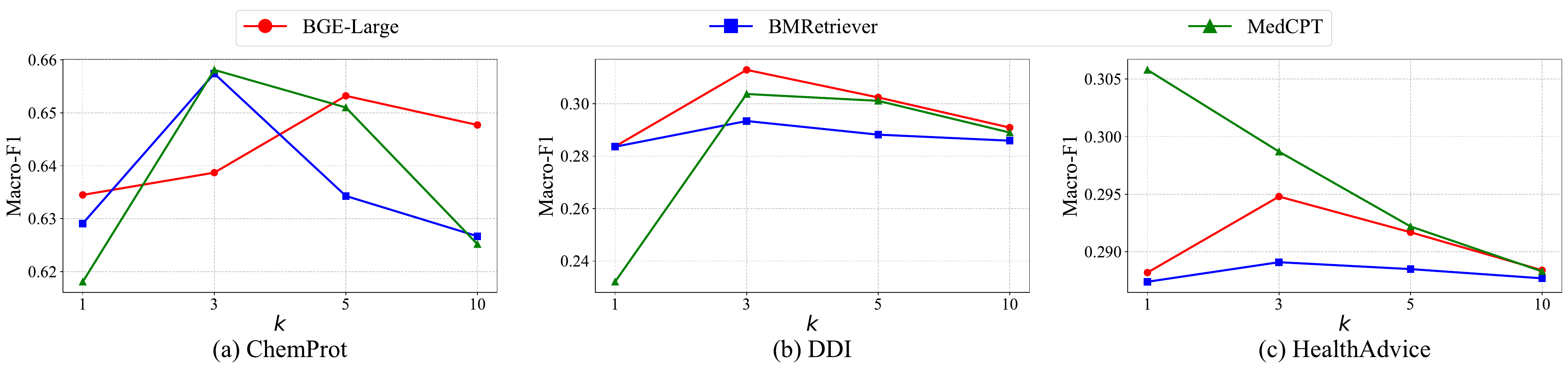}
    \caption{Sensitivity analysis of the budget constraint in the second stage, with Llama 3.1 as the inference LLM.}
    \label{fig_num_llama}
\end{figure}

\subsubsection{On the Effect of Budget Constraint}
Fig.~\ref{fig_num_qwen} and Fig.~\ref{fig_num_llama} illustrate the trend of macro-F1 scores as the budget limit $k$ for the second stage varies.
Overall, Qwen 2.5 consistently outperforms Llama 3.1 across all datasets and retriever models, indicating superior inference capabilities from in-context information for biomedical tasks. This finding aligns with our previous observations.
Regarding specific tasks, the performance improvement is most pronounced on ChemProt (approximately 9.1\% at the optimal $k$) and HealthAdvice (approximately 47.5\%).
Concerning the choice of $k$, increasing its value generally enhances performance across all LLMs and tasks. The optimal macro-F1 performance is typically achieved at $k=3$ or $5$.
However, performance gains diminish beyond $k=3$, becoming marginal. Notably, further increasing $k$ from 5 to 10 often results in negligible or even negative gains across nearly all experiments.
This pattern suggests that the quality of demonstrations is more critical than quantity. An excessive number of examples may potentially impair LLM performance by introducing redundant or irrelevant knowledge.

Furthermore, regarding the fluctuation of macro-F1 scores with varying $k$ across different retriever models, BGE-Large generally exhibits greater robustness compared to BMRetriever and MedCPT. This robustness likely stems from BGE-Large's training on general-domain data, rather than being specialized solely for biomedical contexts.

\subsection{Case Study}

\begin{figure}[ht]
    \centering
    \includegraphics[width=0.99\columnwidth]{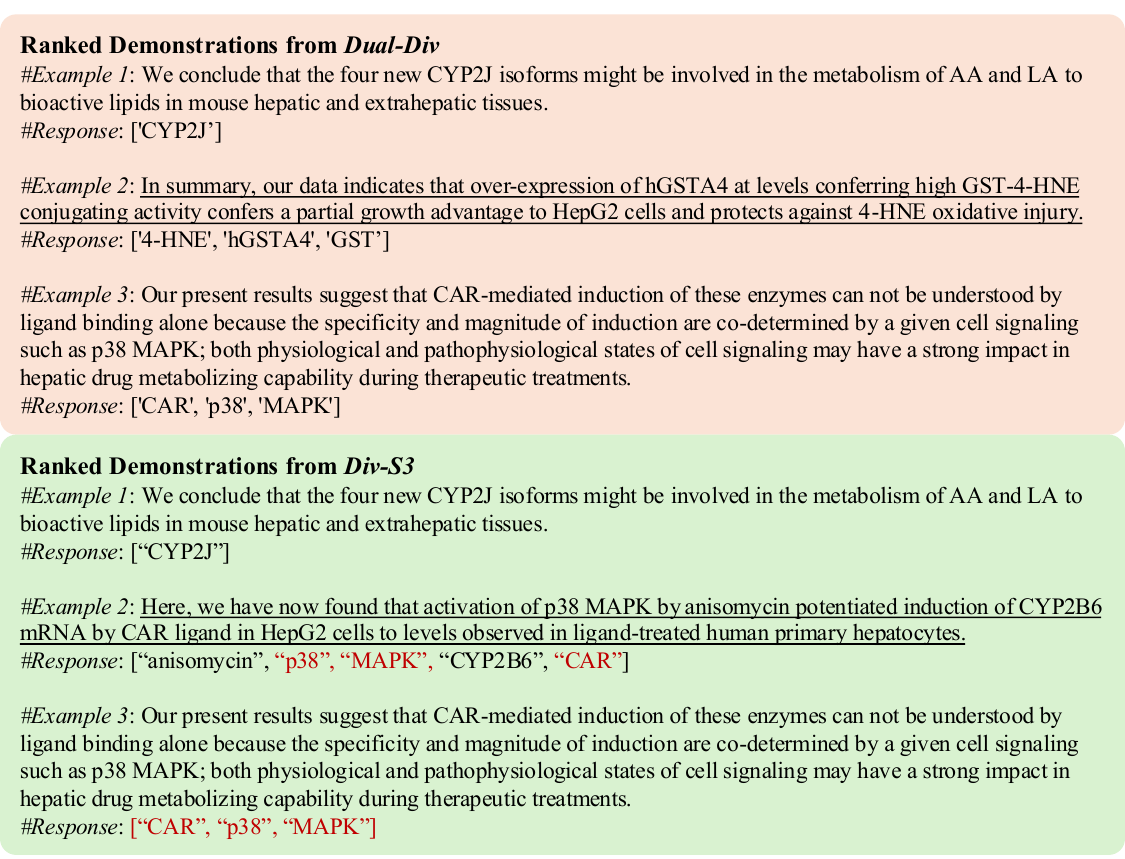}
    \caption{Case study of ranked demonstrations on the ChemProt dataset for NER task from (diversity-enhanced) \textit{Dual-Div} and \textit{Div-S3}, with BGE-Large as the retriever and Qwen 2.5 as the inference LLM.}
    \label{fig_case_ner}
\end{figure}

We further provide case studies to demonstrate the enhanced ability in extracting diverse examples from training corpora. The comparison in Fig.~\ref{fig_case_ner} showcases that the previous Div-S3 method is more likely to select top examples with the same NER tags, while our proposed Dual-Div can greatly alleviate this issue by reducing the similarity within the ranked examples.
See more comparisons regarding the other two tasks in Appendix~\ref{sec_appendix_cases}.

\section{Conclusion}

This paper introduces Dual-Div, a two-stage framework that optimizes demonstration selection for biomedical ICL by jointly maximizing semantic coverage and diversity through submodular optimization. Leveraging a determinantal point process (DPP) for diversity and a facility location function for representativeness, Dual-Div first retrieves a diverse candidate set from large corpora, then ranks examples conditioned on test queries to minimize redundancy. Evaluated across three biomedical NLP tasks (NER, RE, TC) using multiple LLMs (Llama 3.1, Qwen 2.5) and retrievers (BGE-Large, BMRetriever, MedCPT), Dual-Div consistently outperforms baselines—achieving higher macro-F1 scores (up to +5\%) — while proving robust to prompt permutations and class imbalance. Key insights reveal that diversity in initial retrieval (Stage 1) is more critical than ranking (Stage 2), and limiting demonstrations to 3–5 examples optimizes performance, establishing Dual-Div as a data-efficient solution for biomedical ICL.

For future work, we would explore the dynamic adjustment of the trade-off parameter $\lambda$ in Eq.~\eqref{eq_f_S}. In addition, the efficient scaling to extremely large corpora is a practical need, which may be alleviated by developing distributed variants of the lazy greedy algorithm.



\section{Conflict of Interest}
The authors state that they have no conflicts of interest to declare.

\begin{appendices}

\section{Submodularity and Monotonicity of \texorpdfstring{$f(S)$}{f(S)}}\label{app_proof}

\begin{proof}
First of all, $\mathcal{C}(S)$ and $\mathcal{D}(S)$ are both submodular functions. $\mathcal{C}(S)$ is submodular because the linear combinations of submodular functions are still submodular.
$\mathcal{D}(S)$ is submodular because $\mathbf{W}_S$ is positive semi-definite and the determinant of positive semi-definite matrices is log-submodular~\cite{gillenwater2012near}.
It can also be observed that $\mathcal{C}(S)$ is monotonically increasing and $\mathcal{D}(S)$ is monotonically decreasing.
The former holds because if we add a new element $x' \notin S$ into $S$, we have $\forall i \in V$, $\max_{j \in S \cup \{x'\}} w_{i,j} \geq \max_{j \in S} w_{i,j}$. Therefore $\mathcal{C}(S \cup \{x'\}) \geq \mathcal{C}(S)$.
Next, we consider the latter.
Again, $\text{det}(\mathbf{W}_S)$ is non-negative since $\mathbf{W}_S$ is positive semi-definite.
If we add a new element $x' \notin S$ into $S$, we denote $q_S(\bar{\mathbf{e}}_{x'})$ by the orthogonal projection residual of the normalized embedding vector $\bar{\mathbf{e}}_{x'}$ from the retriever model $g(x')$ in terms of the space spanned by vectors $\{\bar{\mathbf{e}}_i\}_{i \in S}$, namely $\bar{\mathbf{e}}_{x'} - \text{proj}_{\text{span}(S)}(\bar{\mathbf{e}}_{x'})$. It holds that $||q_S(\bar{\mathbf{e}}_{x'})||^2 \leq 1$ because $||\bar{\mathbf{e}}_{x'}|| = 1$.
Hence, we have
$
\text{det}(\mathbf{W}_{S \cup \{x'\}}) = \text{det}(\mathbf{W}_S) \cdot ||q_S(\bar{\mathbf{e}}_{x'})||^2 \leq \text{det}(\mathbf{W}_S)
$, leading to $\mathcal{D}(S \cup \{x'\}) \leq \mathcal{D}(S)$.

Intuitively, during the balance of the coverage and diversity of $S$, if we ensure that $\lambda$ is small enough that $0 < \lambda \leq \frac{\mathcal{C}(S \cup \{x'\}) - \mathcal{C}(S)}{\mathcal{D}(S)-\mathcal{D}(S \cup \{x'\})}$ for all subset $S$ and $x' \notin S$, we can obtain a monotonically increasing submodular objective function $f(S) = \mathcal{C}(S) + \lambda \mathcal{D}(S)$.
In our context, if we assume the space is high-dimensional and all vectors in $\mathbf{E}$ are linearly independent, there exists a constant $\varepsilon = 1- \min_{i,j \in V} w_{i,j} > 0$ such that $\mathcal{C}(S \cup \{x'\}) - \mathcal{C}(S) \leq \varepsilon$ for all subset $S$ and $x' \notin S$. Since the upper bound of $\mathcal{D}(S)-\mathcal{D}(S \cup \{x'\})$ is also limited, we can guarantee the existence of the constant $\lambda$.
\end{proof}

\section{Algorithm Details}~\label{sec_appendix_algo}
The pseudocode of our Dual-Div is provided in Algorithm~\ref{algo_dualdiv}.

\begin{algorithm}[ht]
    \caption{The Dual-Div Algorithm}
    \label{algo_dualdiv}
    \textbf{Input}: Corpus set $V$, monotone submodular function $f$, the cardinality limit $k_1$ and $k$ \\
    \textbf{Output}: A subset $T \subseteq S \subseteq V$ with \( |T| \leq k \) and \( |S| \leq k_1 \)
    \begin{algorithmic}[1]
        \Statex \textbf{Phase 1 -- Diversity-enhanced Retrieval}
        \State Initialize \( S \gets \emptyset \) and an empty max-heap \( \mathcal{H} \) \Comment{\( \mathcal{H} \) stores pairs \((x, \delta)\) sorted by \(\delta\)}
        \For{each element \( x \in V \)}
            \State \(\delta_x \gets f(\{x\}) - f(\emptyset)\) \Comment{Initial marginal gain}
            \State \(\mathcal{H}.\textsc{Push}((x, \delta_x))\)
        \EndFor
        \For{\( i = 1 \) to \( k_1 \)}
            \If{\( \mathcal{H} \) is empty}
                \State \textbf{break} \Comment{No elements left to add}
            \EndIf
            \State \((x_{\text{top}}, \delta_{\text{top}}) \gets \mathcal{H}.\textsc{Pop}()\) \Comment{Element with max \(\delta\)}
            \State Compute current marginal gain: \(\delta_{\text{new}} \gets f(S \cup \{x_{\text{top}}\}) - f(S)\)
            \If{\( \mathcal{H} \) is not empty}
                \State \((x_{\text{next}}, \delta_{\text{next}}) \gets \mathcal{H}.\textsc{Peek}()\) \Comment{Next best element in heap}
                \If{\(\delta_{\text{new}} \geq \delta_{\text{next}}\)}
                    \State \( S \gets S \cup \{x_{\text{top}}\} \) \Comment{Add \(x_{\text{top}}\) to solution}
                \Else
                    \State \(\mathcal{H}.\textsc{Push}((x_{\text{top}}, \delta_{\text{new}}))\) \Comment{Reinsert with updated gain}
                \EndIf
            \Else
                \State \( S \gets S \cup \{x_{\text{top}}\} \) \Comment{Add last remaining element}
            \EndIf
        \EndFor
        
        \State \Return $S$

        \Statex
        \Statex \textbf{Phase 2 -- Diversity-enhanced Ranking}
        \State Initialize $T \leftarrow \emptyset$
        \For{$x \in S^*$}
            \State Compute $f( \{x_i\} \mid Q)=f( \{x_i\} \cup Q) - f(Q)$ \Comment{Initial marginal gain}
        \EndFor
        \For{$i=1$ to $k$}
            \State Select $x_{\text{top}} = \argmax _{x \in S^* \setminus T} f( \{x_i\} \mid Q)$
            \State Update $T \leftarrow T\cup \{x_{\text{top}}\}$ \Comment{Add \(x_{\text{top}}\) to solution}
        \EndFor
        \State \Return $T$
    \end{algorithmic}
\end{algorithm}

\section{Implementation Details}~\label{sec_appendix_details}

We set $k_1$ to 100, consistent with established studies, and use $k=3$ as the default value. The impact of $k$ is further analyzed in our sensitivity analysis.

To reduce formatting errors in the inference results generated by LLMs, we found that performing a warm-up procedure using Low-Rank Adaptation (LoRA)~\cite{hu2022lora} between pretraining and ICL inference is beneficial. For LoRA, we set the rank to 64 and alpha to 32.
We fine-tuned the models for 4,000 steps on ChemProt, 2,000 steps on DDI, and 2,000 steps on HealthAdvice. All fine-tuning used the AdamW optimizer with a learning rate of 1e-5. The batch size was 2 during training and 4 during inference across all datasets.

All experiments were conducted on a single NVIDIA A100 GPU with 40GB of memory. Results are reported as the average over five runs.

\section{Prompt Templates}\label{sec_appendix_template}

We present the prompt templates utilized for different NLP tasks below, where the task description is listed in Table~\ref{tab_prompt}.

\begin{lstlisting}[]
    ### Instruction:
    <Task Description>
    
    ### Examples:
    #Input: <Demonstraion#1>
    #Response: <Label#1>
    @$\cdots$@
    #Input: <Demonstraion#@$k$@>
    #Response: <Label#@$k$@>
    
    ### Input:
    <Test Query>
    
    ### Response:
    <Inference Result>
\end{lstlisting}

\begin{table}[ht]
    \centering\small
    \caption{Detailed instructions for each task.}\label{tab_prompt}
    \begin{tabularx}{\textwidth}{l|X}
    \toprule
    Task & \makecell[c]{Description} \\
    \midrule
    \multirow{3}{*}{NER} & Please do a named entity recognition task. You need to accurately recognize chemical or genetic words or terms given the input sentence. The response should only be entities following the form shown in the original sentences. If the given sentence does not include these kinds of entities, just output None. \\
    \midrule
    \multirow{9}{*}{RE} & Please do a relation extraction task. You need to extract the relationship between the given head entity and the tail entity. The response should be in a predefined set: \{`mechanism', `effect, `advice', `int', `none'\}. \\
    & mechanism: this type is used to annotate drug-drug interactions that are described by their pharmacokinetic mechanism.\\
    & effect: this type is used to annotate drug-drug interactions describing an effect or a pharmacodynamic mechanism.\\
    & advice: this type is used when a recommendation or advice regarding a drug interaction is given.\\
    & int: this type is used when a drug-durg interaction appears in the text without providing any additional information.\\
    & none: there are no drug-drug interactions. \\
    \midrule
    \multirow{2}{*}{TC} & Please do a classification task. You need to classify what type advice the input sentence is. The response should be in pre-defined set: (`no', `weak', `strong'). \\
    \bottomrule
    \end{tabularx}
\end{table}

\section{More Experimental Results}

\begin{table}[ht]
    \centering
    \caption{ICL Performance on different NLP datasets using Llama 3.1 (8B) and Qwen 2.5 (7B) as inference LLMs and BMRetriever as the retriever.}\label{tab_main_bmretriever}
    \begin{tabular}{clcccccc}
    
    \toprule
    \multirow{2}{*}{LLM} & \multicolumn{1}{c}{\multirow{2}{*}{Method}} & \multicolumn{2}{c}{ChemProt (NER)} & \multicolumn{2}{c}{DDI (RE)} & \multicolumn{2}{c}{HealthAdvice (TC)} \\
    \cmidrule(lr){3-4} \cmidrule(lr){5-6} \cmidrule(lr){7-8}
    & & acc. & macro-F1 & acc. & macro-F1 & acc. & macro-F1 \\
    \midrule
    \multirow{5}{*}{Qwen 2.5} & Random-Simliar & 0.7118 & 0.6673 & 0.7499 & 0.2513 & 0.7592 & 0.2921 \\
    & Div-S3 & 0.7028 & 0.6881 & 0.7065 & 0.2689 & 0.7644 & 0.3350 \\
    & Div*-S3 & \textbf{0.7364} & 0.7220 & \textbf{0.7756} & 0.3049 & \textbf{0.7667} & \textbf{0.3456} \\
    & Div-S3* & 0.7085 & 0.7051 & 0.7065 & 0.2689 & 0.7656 & 0.3390 \\
    & Dual-Div & 0.7324 & \textbf{0.7261} & 0.7716 & \textbf{0.3180} & \textbf{0.7667} & \textbf{0.3456} \\
    \midrule
    \multirow{5}{*}{Llama 3.1} & Random-Simliar & 0.6815 & 0.6261 & 0.6678 & 0.2494 & 0.6751 & 0.2786 \\
    & Div-S3 & 0.6926 & 0.6343 & 0.6797 & 0.2582 & 0.6897 & 0.2830 \\
    & Div*-S3 & 0.7082 & 0.6493 & \textbf{0.6950} & 0.2844 & \textbf{0.7615} & 0.2874 \\
    & Div-S3* & 0.7010 & 0.6521 & 0.6813 & 0.2862 & 0.7177 & 0.2883 \\
    & Dual-Div & \textbf{0.7095} & \textbf{0.6574} & 0.6936 & \textbf{0.2932} & 0.7563 & \textbf{0.2890} \\
    \bottomrule
    \end{tabular}
\end{table}

\begin{table}[ht]
    \centering
    \caption{ICL Performance on different NLP datasets using Llama 3.1 (8B) and Qwen 2.5 (7B) as inference LLMs and MedCPT as the retriever.}\label{tab_main_medcpt}
    \begin{tabular}{clcccccc}
    
    \toprule
    \multirow{2}{*}{LLM} & \multicolumn{1}{c}{\multirow{2}{*}{Method}} & \multicolumn{2}{c}{ChemProt (NER)} & \multicolumn{2}{c}{DDI (RE)} & \multicolumn{2}{c}{HealthAdvice (TC)} \\
    \cmidrule(lr){3-4} \cmidrule(lr){5-6} \cmidrule(lr){7-8}
    & & acc. & macro-F1 & acc. & macro-F1 & acc. & macro-F1 \\
    \midrule
    \multirow{5}{*}{Qwen 2.5} & Random-Simliar & 0.6864 & 0.6704 & 0.7559 & 0.2488 & 0.7446 & 0.2796 \\
    & Div-S3 & \textbf{0.7421} & 0.7064 & 0.7559 & 0.2789 & 0.7615 & 0.2923 \\
    & Div*-S3 & \textbf{0.7421} & 0.7064 & \textbf{0.8070} & 0.2588 & 0.7615 & 0.2963 \\
    & Div-S3* & 0.7190 & 0.7119 & 0.7771 & 0.3005 & \textbf{0.7620} & 0.3039 \\
    & Dual-Div & 0.7284 & \textbf{0.7155} & 0.7870 & \textbf{0.3115} & 0.7598 & \textbf{0.3179} \\
    \midrule
    \multirow{5}{*}{Llama 3.1} & Random-Simliar & 0.6351 & 0.5184 & 0.6966 & 0.2676 & 0.6230 & 0.2805 \\
    & Div-S3 & 0.6681 & 0.6225 & \textbf{0.7440} & 0.2839 & 0.7158 & 0.2917 \\
    & Div*-S3 & \textbf{0.6695} & 0.6228 & 0.7089 & 0.2983 & 0.7085 & 0.2954 \\
    & Div-S3* & 0.6444 & 0.6510 & 0.7431 & 0.2898 & \textbf{0.7558} & 0.2889 \\
    & Dual-Div & 0.6507 & \textbf{0.6581} & 0.7341 & \textbf{0.3039} & 0.7298 & \textbf{0.2988} \\
    \bottomrule
    \end{tabular}
\end{table}

\subsection{ICL Performance}~\label{sec_appendix_main_exp}
We supplement the in-context learning results of all methods for different tasks with BMRetriever and MedCPT as the retriever models in Table~\ref{tab_main_bmretriever} and Table~\ref{tab_main_medcpt} separately.

\begin{figure}[ht]
    \centering
    \includegraphics[width=0.98\columnwidth]{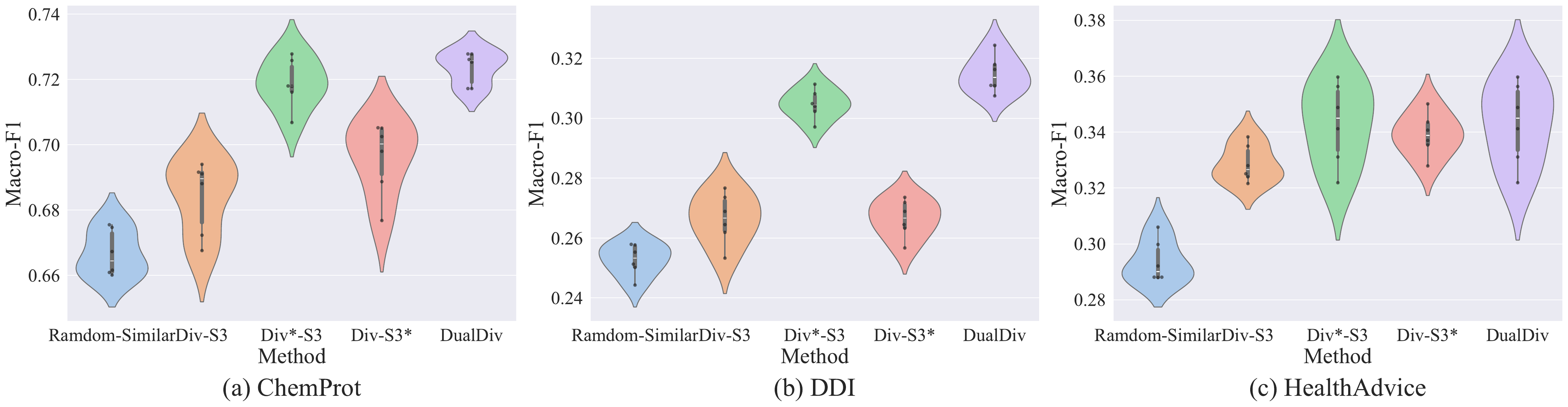}
    \caption{Sensitivity analysis regarding the permutation ordering in the LLM's prompts of the final demonstration set. Given BMRetriever as the retriever model, for each set of 3 demonstrations from different methods, we plot the violin graphics of macro-F1 scores of all 6 possible permutations.}
    \label{fig_permute_bm}
\end{figure}

\begin{figure}[ht]
    \centering
    \includegraphics[width=0.98\columnwidth]{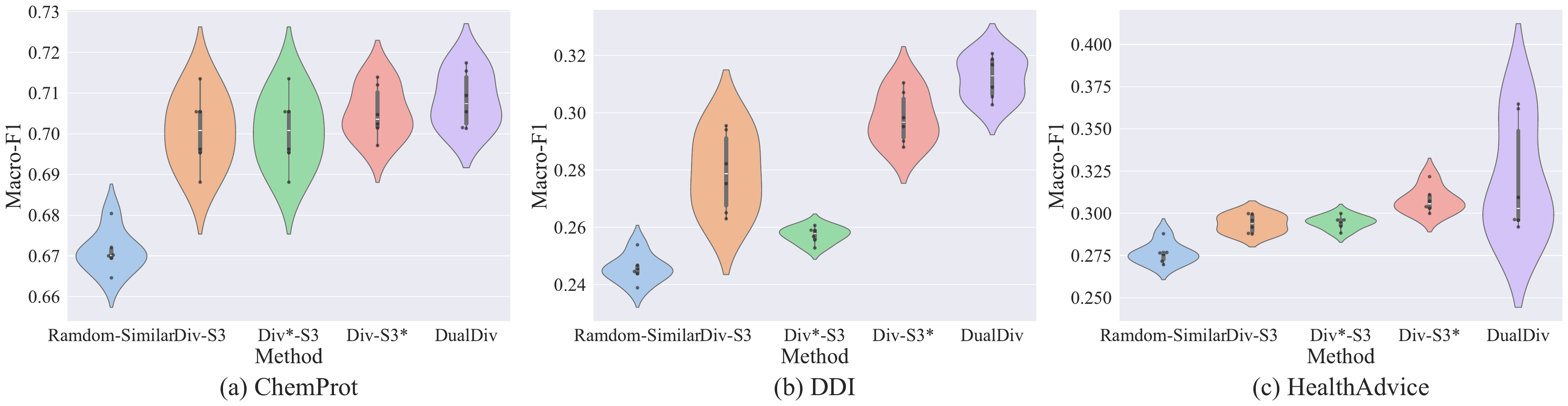}
    \caption{Sensitivity analysis regarding the permutation ordering in the LLM's prompts of the final demonstration set. Given MedCPT as the retriever model, for each set of 3 demonstrations from different methods, we plot the violin graphics of macro-F1 scores of all 6 possible permutations.}
    \label{fig_permute_medcpt}
\end{figure}

\subsection{Effect of Example Ordering}~\label{sec_appendix_order}
For the sensitivity analysis regarding the permutation ordering within LLM's input prompts, we supplement the violin plots of all possible permutations of the examples in the final set $T$, using BMRetriever and MedCPT as the retriever, in Fig.~\ref{fig_permute_bm} and Fig.~\ref{fig_permute_medcpt} separately.

\begin{figure}[ht]
    \centering
    \includegraphics[width=0.99\columnwidth]{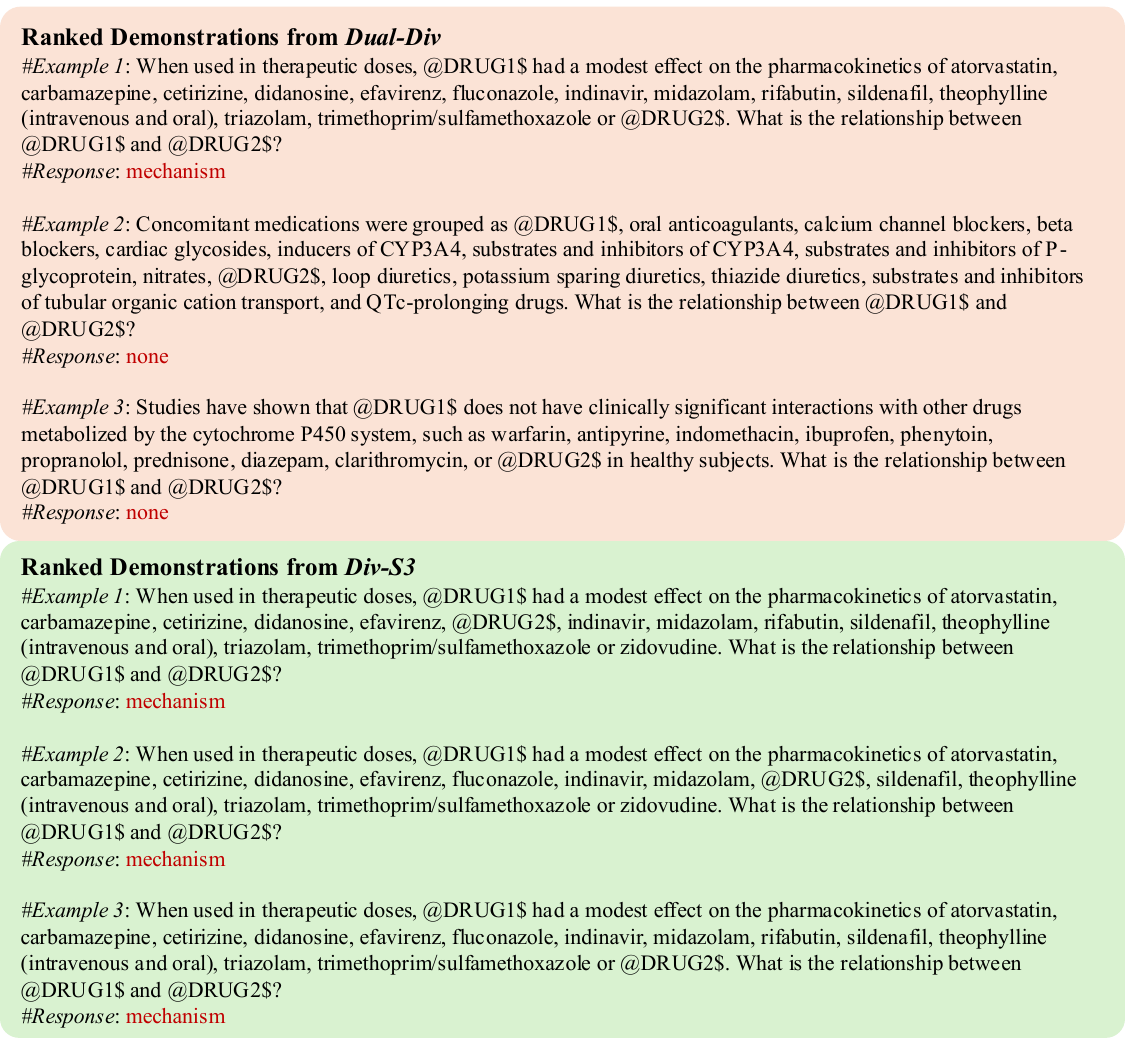}
    \caption{Case study of ranked demonstrations on the DDI dataset for RE task from (diversity-enhanced) \textit{Dual-Div} and \textit{Div-S3}, with BGE-Large as the retriever and Qwen 2.5 as the inference LLM.}
    \label{fig_case_re}
\end{figure}

\begin{figure}[ht]
    \centering
    \includegraphics[width=0.99\columnwidth]{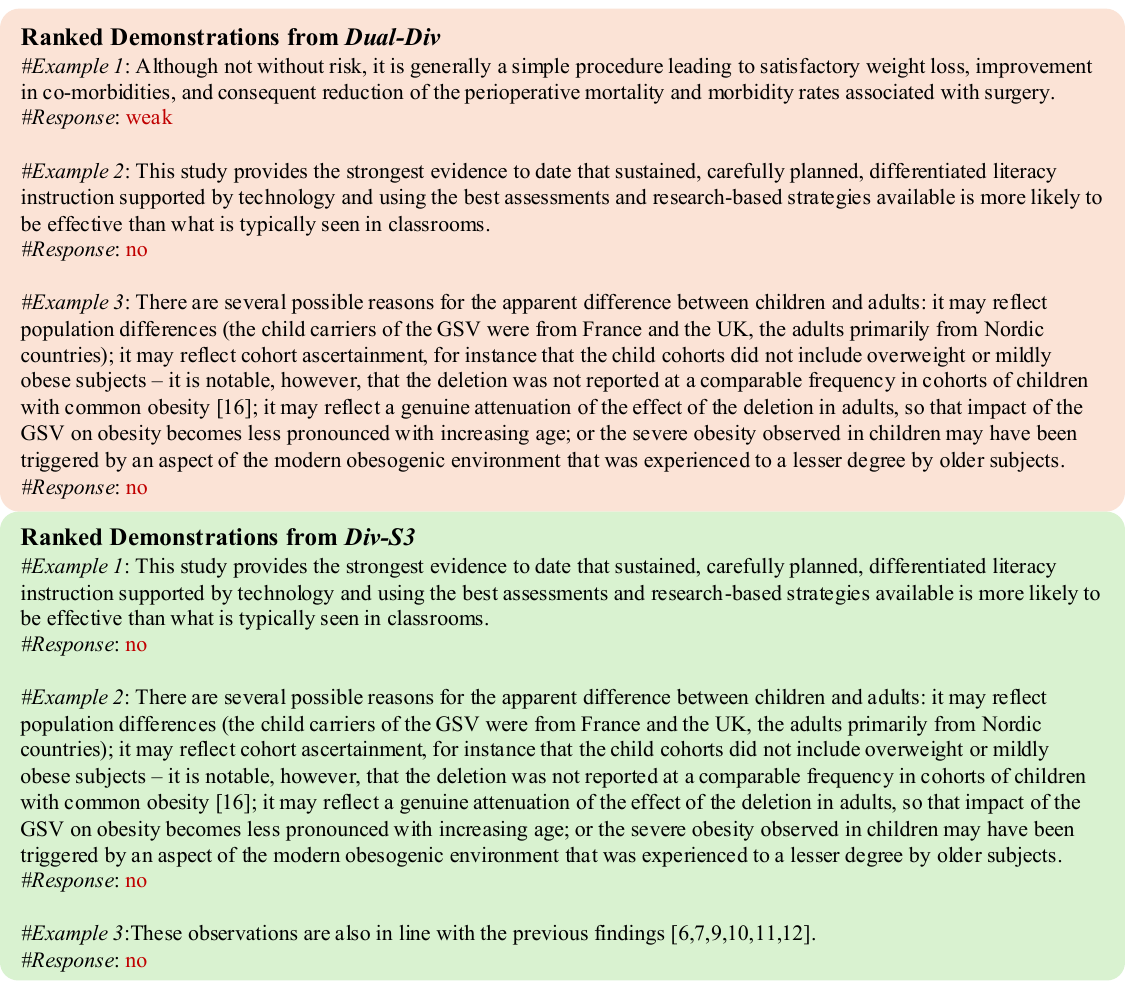}
    \caption{Case study of ranked demonstrations on the HealthAdvice dataset for TC task from (diversity-enhanced) \textit{Dual-Div} and \textit{Div-S3}, with BGE-Large as the retriever and Qwen 2.5 as the inference LLM.}
    \label{fig_case_tc}
\end{figure}

\subsection{Case Study}~\label{sec_appendix_cases}
Finally, we present case studies: the top demonstrations selected by Dual-Div and Div-S3 on the DDI and HealthAdvice datasets, shown in Fig.~\ref{fig_case_re} and Fig.~\ref{fig_case_tc}, respectively.

\end{appendices}

\clearpage

\bibliographystyle{plain}
\bibliography{references}


\end{document}